\documentclass[sigconf]{acmart}

\AtBeginDocument{%
  \providecommand\BibTeX{{%
    \normalfont B\kern-0.5em{\scshape i\kern-0.25em b}\kern-0.8em\TeX}}}

\setcopyright{acmcopyright}
\copyrightyear{2020}
\acmYear{2020}
\acmDOI{10.1145/3394171.3413826}
\acmConference[MM '20]{Proceedings of the 28th ACM
International Conference on Multimedia}{October 12--16, 2020}{Seattle, WA, USA}
\acmBooktitle {Proceeding of the 28th ACM International Conference on Multimedia
(MM '20), October 12--16, 2020, Seattle, WA, USA}
\acmPrice{15.00}
\acmISBN{978-1-4503-7988-5/20/10}

\usepackage{array}
\newcommand{\PreserveBackslash}[1]{\let\temp=\\#1\let\\=\temp}
\newcolumntype{C}[1]{>{\PreserveBackslash\centering}p{#1}}
\newcolumntype{R}[1]{>{\PreserveBackslash\raggedleft}p{#1}}
\newcolumntype{L}[1]{>{\PreserveBackslash\raggedright}p{#1}}
\usepackage{multirow}

\begin{document}
\fancyhead{}
\title{KBGN: Knowledge-Bridge Graph Network for Adaptive Vision-Text Reasoning in Visual Dialogue}

\author{Xiaoze Jiang$^1$, Siyi Du$^{1}$, Zengchang Qin$^{1,2*}$, Yajing Sun$^{3}$, Jing Yu$^{3}$}\affiliation{$^1$Intelligent Computing and Machine Learning Lab, School of ASEE, Beihang University, Beijing, China\\$^2$AI Research, Codemao Inc.\quad $^3$Institute of Information Engineering, Chinese Academy of Sciences, Beijing, China \\ \{xzjiang, siyi\_wind, zcqin\}@buaa.edu.cn, \{sunyajing, yujing02\}@iie.ac.cn }
\begin{abstract}
  Visual dialogue is a challenging task that needs to extract implicit information from both visual (image) and textual (dialogue history) contexts. 
  Classical approaches pay more attention to the integration of the current question, vision knowledge and text knowledge, despising the heterogeneous semantic gaps between the cross-modal information. In the meantime, the concatenation operation has become de-facto standard to the cross-modal information fusion, which has a limited ability in information retrieval. In this paper, we propose a novel Knowledge-Bridge Graph Network (KBGN) model by using graph to bridge the cross-modal semantic relations between vision and text knowledge in fine granularity, as well as retrieving required knowledge via an adaptive information selection mode. 
  %KBGN can capture cross-modal information through a series of consecutive modules: Knowledge Encoding, Knowledge Storage and Knowledge Retrieval. 
  %Specifically, Knowledge Encoding helps grip the intra-modal information, including intra-modal entities and their visual or context dependence within the intra-modal knowledge. Knowledge Storage stores information by bridging relation between inter-modal knowledge via designing cross-modal graph neural networks. Knowledge Retrieval maintains a more flexible information selection mode by retrieving knowledge adaptively. 
  Moreover, the reasoning clues for visual dialogue can be clearly drawn from intra-modal entities and inter-modal bridges. Experimental results on \emph{VisDial v1.0} and \emph{VisDial-Q} datasets demonstrate that our model outperforms existing models with state-of-the-art results. %on the top of such framework,
\end{abstract}



\iffalse
\begin{CCSXML}
\iffalse
<ccs2012>
 <concept>
  <concept_id>10010520.10010553.10010562</concept_id>
  <concept_desc>Computer systems organization~Embedded systems</concept_desc>
  <concept_significance>500</concept_significance>
 </concept>
 <concept>
  <concept_id>10010520.10010575.10010755</concept_id>
  <concept_desc>Computer systems organization~Redundancy</concept_desc>
  <concept_significance>300</concept_significance>
 </concept>
 <concept>
  <concept_id>10010520.10010553.10010554</concept_id>
  <concept_desc>Computer systems organization~Robotics</concept_desc>
  <concept_significance>100</concept_significance>
 </concept>
 <concept>
  <concept_id>10003033.10003083.10003095</concept_id>
  <concept_desc>Networks~Network reliability</concept_desc>
  <concept_significance>100</concept_significance>
 </concept>
</ccs2012>
\fi
<ccs2012>
   <concept>
       <concept_id>10010147.10010178.10010224.10010225.10010231</concept_id>
       <concept_desc>Computing methodologies~Visual content-based indexing and retrieval</concept_desc>
       <concept_significance>500</concept_significance>
       </concept>
 </ccs2012>
\end{CCSXML}

\fi
% Adaptive Information Retrieval
\iffalse
\begin{teaserfigure}
  \includegraphics[width=\textwidth]{sampleteaser}
  \caption{Seattle Mariners at Spring Training, 2010.}
  \Description{Enjoying the baseball game from the third-base
  seats. Ichiro Suzuki preparing to bat.}
  \label{fig:teaser}
\end{teaserfigure}
\fi

\maketitle

\section{Introduction}
\label{Introduction}

Thanks to the development of both natural language processing and computer vision research, it inspired a surge of interest in integrating vision and language to construct a more general intelligent agent dealing with applications ranging from Image Captioning \cite{johnson2016densecap,wu2016value}, Referring Expressions  \cite{Wang2019Neighbourhood}, Visual Question Answering (VQA) \cite{Agrawal2017VQA} to Visual Dialogue \cite{Das2017Visual}. Different from the VQA task, visual dialogue is an on-going conversation about the image, and the relations among visual objects are dynamically shifted with conversational contexts. 
This requires the agent to retrieve information from both vision knowledge (image) and text knowledge (dialogue history) \cite{Das2017Visual}. Between these two modalities, there exists a heterogeneous semantic gap of implicit referring relations cross modalities. How can we effectively model this semantic gap is a key challenge in visual dialogue task. 
% , which means transferring from one modality to another has become de-facto standard in Visual Dialogue.
%To bridge the cross-modal gap in visual dialogue, 
\begin{figure}[t]
\centering
\includegraphics[width=8cm]{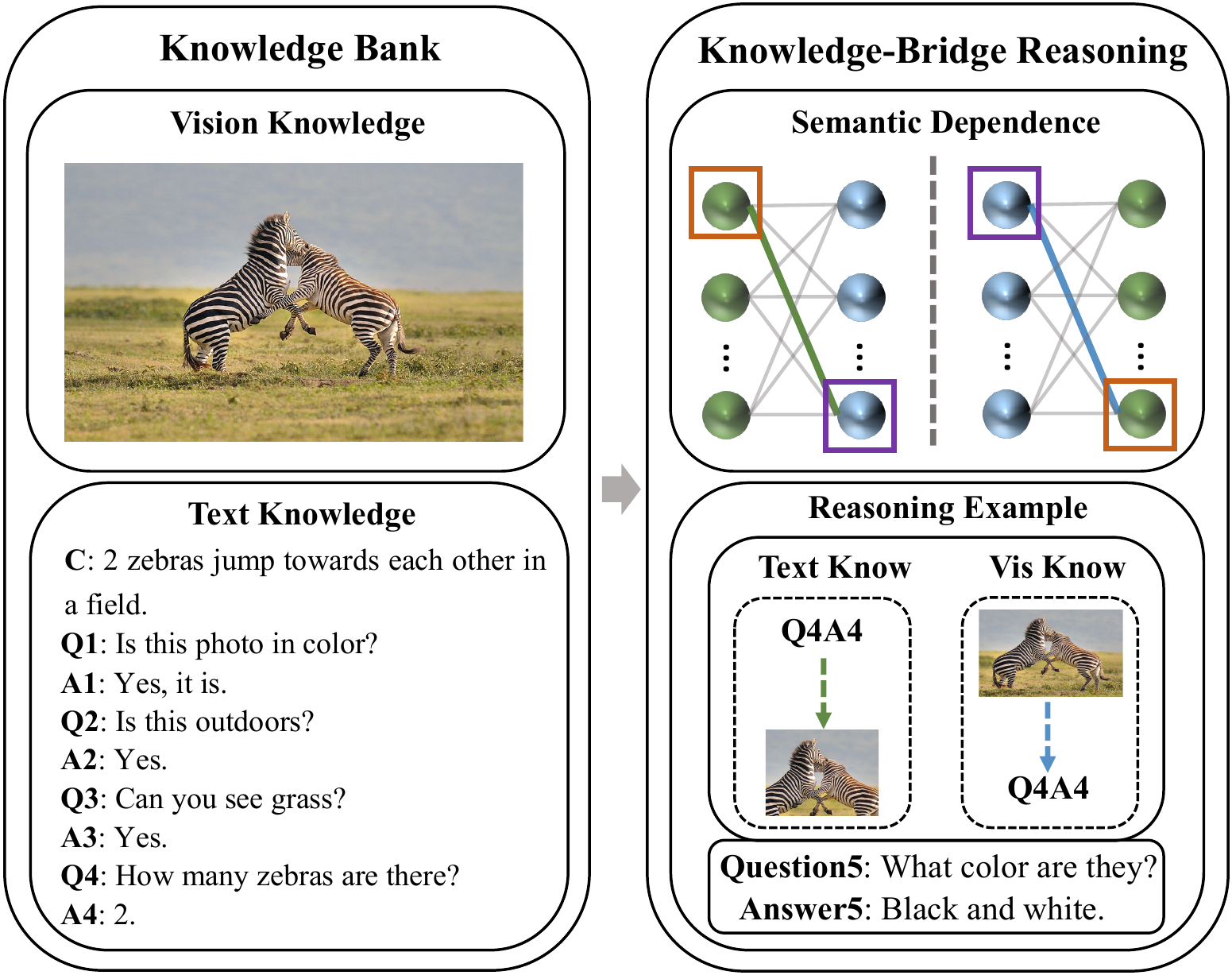}
\caption{A schematic illustration of KBGN. Given inputs of the image and the dialogue history, the underlying semantic dependence (text to vision dependence: blue lines, vision to text dependence: green lines) is captured by the model in local granularity: Q-A pairs (green balls) and visual objects (blue balls). The key information for answering a given question can be retrieved in cross-modal knowledge bank.
%visual objects and Q-A pairs. Green balls: Q-A pairs, blue balls: visual objects, green
%for answering Q5, adaptive information are retrieved related to QA pairs of Q3 and Q4. %Green (blue) lines represent cross-modal bridge. Green (blue) boxes represent the information from text (vision) knowledge.
}
\label{basicIdea}
\end{figure}

In order to solve this challenging problem, some previous studies \cite{wu2018areyou,yang2019making} tried to fuse multiple features of vision and text knowledge. Some other works focused on solving the visual reference resolution problem \cite{seo2017visual,KotturSatwik2018Visual}, they basically fuse the vision and text information through attention mechanism to capture key frames that are the most relevant to the current question. However, these works merely assign the global representation from one modality to another, i.e. encode the whole vision information (or text information) to a vector to attend each element from text information (or vision information). As a result, they fail in bridging the cross-modal semantic gap in fine granularity, which means that 
they cannot capture the underlying semantic dependence between each entity in two modalities. It cannot be neglected as the vision knowledge is visual object-relationship-sensitive \cite{jiang2019daulvd,guo2020iterative} and the text knowledge is context-sensitive \cite{Zheng2019Reasoning}. We thereby take these underlying semantic dependence in visual and textual contexts into consideration to bridge the vision-text semantic gap, which means to relate visual objects and question-answering (Q-A) pairs (Figure \ref{basicIdea}). 
In intra-modal relational reasoning, the graph structure is used to capture the vision or text dependence. For inter-modal relational modeling, the cross-modal graph network is employed to exploit the latent dependence between each entity of vision and text knowledge. For example, as shown in Figure \ref{basicIdea}: given `` \emph{Q5: What color are they?}'', the agent first targets on \emph{Q4A4} according to the textual dependence, for \emph{Q4A4} illustrates  that `` \emph{they}'' refers to `` \emph{2 zebras}''. If we can bridge semantic information between each entity of text and vision knowledge, the inter-modal dependence can be captured in local granularity. Thus our model can accurately figure out the representation of \emph{Q4A4} in vision domain instantly, and then extracts the color attribute for `` \emph{2 zebras}'' in vision knowledge. 
%Symmetrically, the cross-modal bridge can also help vision knowledge map visual contents to text domain.

%retrieves information in text knowledge and then targets on \emph{Q4A4}, for \emph{Q4A4} illustrates  that `` \emph{they}'' refers to `` \emph{2 zebras}''. Since the bridge exists between text knowledge and vision knowledge, our model can accurately figure out the representation of \emph{Q4A4} in vision domain immediately, and then extracts the color attribute for `` \emph{2 zebras}'' in vision knowledge. Symmetrically, the cross-modal bridge can also help vision knowledge map visual contents to text domain.

%between each local entities of vision and text knowledge

Even if the cross-modal relation is well modeled, there still exists another problem: how to retrieve the key information for answer prediction. Classical methods adopt a rigid information selection mode via directly concatenating cross-modal information \cite{seo2017visual, wu2018areyou, kang2019dual,Das2017Visual,guo2020iterative,KotturSatwik2018Visual,jiang2019daulvd}. This may introduce some redundant information because the information from different modality has different contributions to answering a question. Distinct from previous work, we propose an adaptive information retrieval mode to capture reasoning clues. As shown in Figure \ref{basicIdea}: When answering `` \emph{Q3: Can you see grass?}'', the agent takes more vision information. While answering `` \emph{Q4: How many zebras are there?}'', the agent pays more attention to text knowledge (i.e. from Caption \emph{C}: ``2 zebras''). By this way, the most relevant information in different modalities can be selected to predict the answer.

In this paper, we propose the Knowledge-Bridge Graph Network (KBGN) to model the cross-modal semantic gap and retrieve question-related information adaptively in visual dialogue. KBGN utilizes the current question as the query to guide and search relevant information from both vision and text knowledge. The framework contains three main modules, i.e. \emph{Knowledge Encoding}, \emph{Knowledge Storage} and \emph{Knowledge Retrieval}. Specifically, the Knowledge Encoding module encodes the visual and textual information from inputs to capture the text and the vision relations within the intra-modal knowledge. The Knowledge Storage module further bridges the vision and text gap to store the text-riched vision information and the vision-riched text information to the \emph{knowledge bank}, respectively. The Knowledge Retrieval module adaptively selects relative information from vision and text knowledge for the final answer. 

Our main contributions can be summarized as follows: (1) We propose a novel framework using graph network models to bridge the cross-modal knowledge information and capture the implicit dependence in two modalities. Furthermore, the model follows a more flexible information selection mode, which can adaptively retrieve information from both vision and text knowledge. (2) To our best knowledge, we are the first to apply graph structure to build cross-modal information bridges for capturing the underlying dependence between vision and text modalities in fine granularity in visual dialogue. (3) The proposed model achieves the state-of-the-art results on two large-scale datasets: VisDial v1.0 \cite{Das2017Visual} and VisDial-Q \cite{ujain2018two}. More importantly, the reasoning clues can be clearly seen by using the proposed model. 

\section{Related work}

% \subsection{VQA}

\textbf{Visual Dialogue} \cite{Das2017Visual} is a challenging task in vision and language problems, it requires to consider multi-round dialogue history and the image in order to find the best candidate answer. Classical methods studied the fusion of the current question, dialogue history and image via using attention mechanism \cite{seo2017visual, wu2018areyou, kang2019dual,Das2017Visual}. For example, \citeauthor{seo2017visual} \cite{seo2017visual} leveraged a neural attention network to resolve the current reference 
problem in visual dialogue. \citeauthor{wu2018areyou} \cite{wu2018areyou} employed a sequential co-attention model to selectively focus on images and dialogue history. The general idea of these works is assigning the global representation from one modality to attend the other and then adopting Late Fusion \cite{Das2017Visual} to incorporate information.  However, this may cause two problems: 1) Semantic gaps between the cross-modal information cannot be well captured by using global information to attend local information from another modality. 2) It's unreasonable to view each piece of information equally and utilize the concatenation operation when answering diverse questions. To solve the above problems, we consider 
to use graph neural network to model cross-modal relations in fine granularity and further design a novel model to retrieve information adaptively based on the query.

%图片
\begin{figure*}[t]
\centering
\includegraphics[width=18cm]{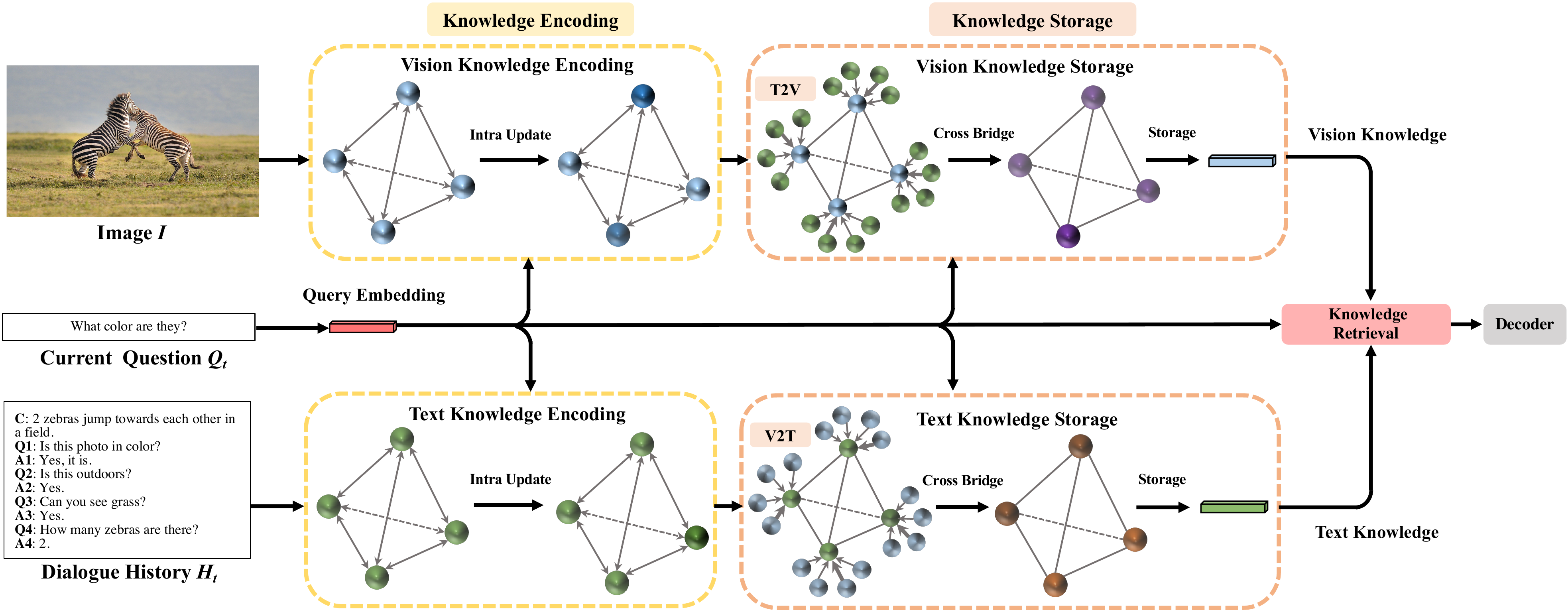}
\caption{Overall structure of the KBGN model, where T2V is the Text to Vision GNN and V2T is the Vision to Text GNN. The model mainly contains three main modules in each modality: \emph{Knowledge Encoding}, \emph{Knowledge Storage} and \emph{Knowledge Retrieval}. }
\label{model_pic}
\end{figure*}

\noindent \textbf{Graph Neural Network (GNN)} was first introduced by \citeauthor{gori2005new} \cite{gori2005new}, and further studied in \cite{scarselli2008graph}. Research in GNN has become a hot topic in deep learning, and it can be roughly categorized into graph convolution network \cite{defferrard2016convolutional,kipf2017semi,monti2017geometric}, graph attention network \cite{velivckovic2018graph,lee2018graph}, graph auto-encoder \cite{cao2016deep,wang2016structural}, graph generation network \cite{bojchevski2018netgan,de2018molgan} and graph spatial-temporal network \cite{yu2018spatio,yan2018spatial}. Due to the brain-like reasoning process and a more graceful, concise way to present dependence information, GNN has been used in some studies of vision and language tasks \cite{li2019relation,teney2017graph}. For visual dialogue, there were some graph-based works that focused on the dialogue structure recovering \cite{Zheng2019Reasoning}, deep visual understanding \cite{jiang2019daulvd,guo2020iterative} and answer information revolving \cite{schwartz2019factor}. Although they have achieved a significant improvement in performance, the core idea of modeling cross-modal semantic gap is still the same 
with previous approaches.
The application of GNN to capture semantic dependence between cross-modal information in visual dialogue has been less studied, though it has a great performance in capturing the structured data  \cite{Zheng2019Reasoning,jiang2019daulvd}. For the first time, we propose to use the graph edge to construct bridges between nodes of vision (image) and text knowledge (dialogue history) in visual dialogue, it helps to shrink the cross-modal semantic gap between two domains of knowledge.

\section{Methodology}

According to \citeauthor{Das2017Visual} \cite{Das2017Visual}, 
in the visual dialogue task, an agent is required to reason based on a given image $I$, caption $C$, dialogue history $H_t = \{ C, (Q_1,A_1),...,(Q_{t-1},A_{t-1})\}$, and the current question $Q_t$ at round $t$. The task is to rank a list of 100 candidate answers $\mathbb{A}  = \{ A_1, A_2,...,A_{100}\}$ and return the best answer $A_t$ to $Q_t$. %Our model is built on the basic late fusion (LF) framework \cite{Das2017Visual}.  
As shown in Figure \ref{model_pic}, the model consists of three modules: \emph{Knowledge Encoding}, \emph{Knowledge Storage} and \emph{Knowledge Retrieval}.  Each module performs its own function but cooperates with each other. Overall speaking, our model first encodes diverse information separately, then stores cross-modal information with underlying relation and finally retrieves required information when the question is asked. Specifically, given the current question $Q_t$, the knowledge $I$ and knowledge $H_t$, we first construct an intra-modal GNN to capture the intra-modal semantic dependence via Knowledge Encoding module. To model the semantic gap between the vision and text knowledge, we use cross-modal GNN to grip the inter-modal semantics between the intra-dependence-aware knowledge in local granularity. Then global knowledge information is stored into knowledge bank under the comprehensive consideration of intra-dependence-aware and inter-dependence-aware local knowledge via Knowledge Storage module. In the end, the Knowledge Retrieval module adopts a more adaptive selection strategy to retrieve the key knowledge from vision and text knowledge bank adaptively for the answer prediction. The reasoning clues are from local entities of the intra-modal knowledge and bridges of the inter-modal connection. We will introduce the details of each module in this section.

\subsection{Knowledge Encoding}

Different from the textual information in other NLP tasks, such as document summarization \cite{Luhn1958theauto,tan2017abstractive,sun2019feasibility}, the textual information in visual dialogue has obviously structured characteristics between each Q-A pair \cite{Zheng2019Reasoning}. In the meantime, distinct from other vision-language tasks, like VQA, the relationship between each visual entity is widely asked \cite{jiang2019daulvd}. Take these two key problems into consideration and inspired by the work of \cite{Zheng2019Reasoning} and \cite{jiang2019daulvd}, we encode vision and text knowledge via graph neural network to further select the most important information within intra-modality. More importantly, the graph structure is convenient to build a bridge between vision and text knowledge, for the nodes in the graph can be connected by the edges naturally. Knowledge Encoding module is presented with yellow boxes in Figure \ref{model_pic}, and contains two parts: \emph{Vision Knowledge Encoding} and \emph{Text Knowledge Encoding}.

\noindent \textbf{Vision Knowledge Encoding}  

\begin{figure}
\centering
\includegraphics[width=8.2cm]{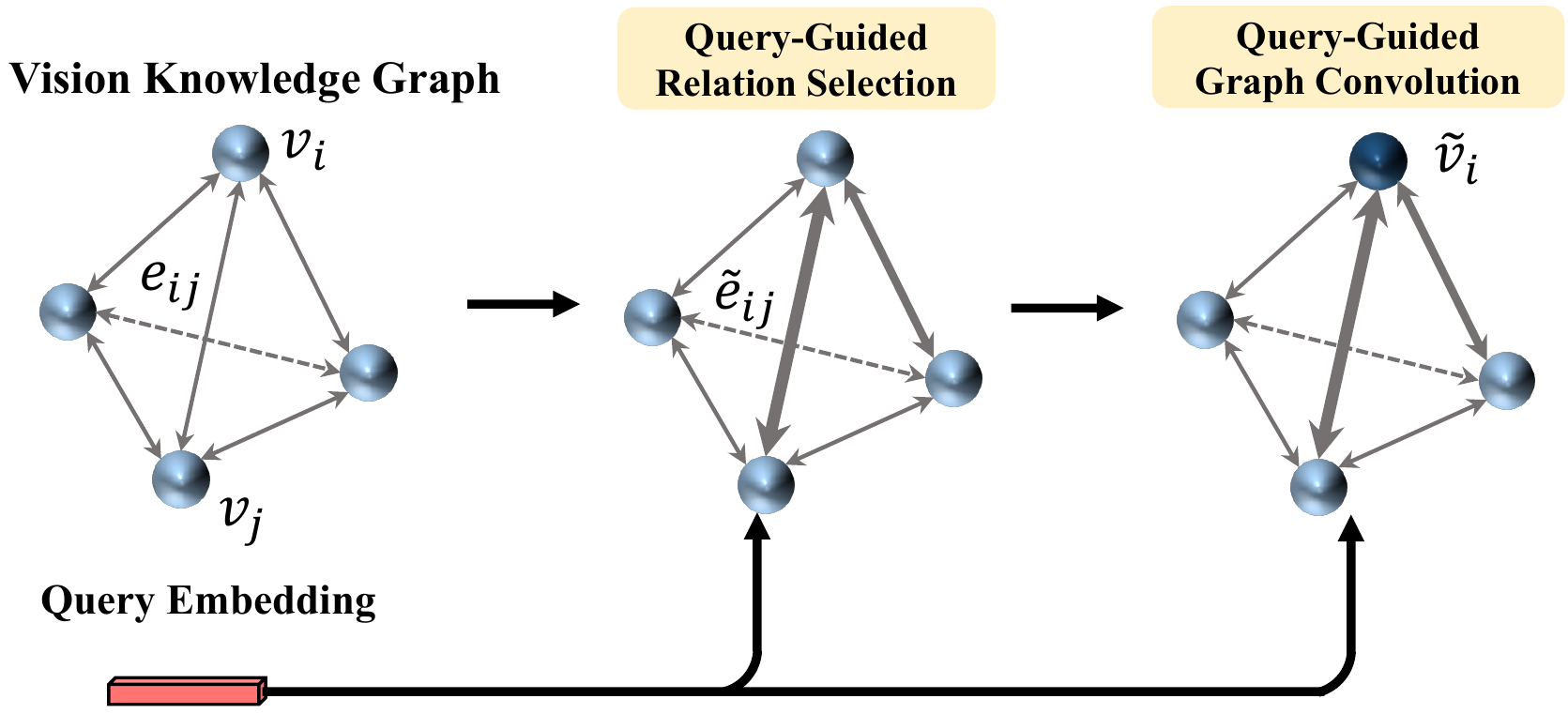}
\caption{The process of Intra Update has two steps:  \emph{Query-guided relation selection} and \emph{query-guided graph convolution}. }
\label{inter-update}
\end{figure}

\noindent For the vision knowledge graph, the nodes $V=\{v_i\}^{N}$ are visual entities, which can be detected by a pre-trained Faster-RCNN \cite{Ren2017Faster} where $N$ is the number of detected objects. The edges $E=\{e_{ij}\}^{N\times N}$ are the visual relationships between nodes provided by a visual relationship encoder in \cite{zhang2019large}. The vision knowledge graph is a fully-connected graph by assuming that the relationship exists in each node and the unknown-relationship is treated as a special relationship. Based on the vision knowledge graph, we further abandon the useless information and select query-relevant information for the Knowledge Storage module. As shown in Figure \ref{inter-update}, the vision knowledge graph is updated as follows.
% The edges that $E=\{e_{ij}\}^{N\times N}$ are the visual relationships between nodes, which are provided by a visual relationship encoder \cite{zhang2019large}. 

\emph{Query-Guided Relation Selection}: Under the guidance of current query vector $Q_t$ (encoded by an unbidirectional LSTM \cite{hochreiter1997long}), the query-relevant relation information is selected by the following operations:
\begin{align}
\alpha _{ij} = softmax(\textbf{W}_e  (\textbf{W}_1   Q_{t}  \circ \textbf{W}_2  e_{ij}))&\\
\widetilde{e}_{ij} = \alpha _{ij}  e_{ij}&
\end{align}
where ``$\circ$'' denotes the element-wise product, $\alpha _{ij}$ (as well as $\beta _{ij}, \gamma _{ij},$ $\delta_{ij}, \eta_i^v, \mu_i^v$) is attention value and $\textbf{W}_1$ (as well as $\textbf{W}_2, ...,$ $ \textbf{W}_{11}$) is the linear transformation layer$\footnote{For conciseness, all the bias terms of linear transformations in this paper are omitted.}$. %For conciseness, all the bias terms of linear transformations in this paper are omitted. 

\emph{Query-Guided Graph Convolution}: Each node in the graph is updated by its neighborhoods and relevant relationships:
\begin{align}
    \beta _{ij} = softmax(\textbf{W}_{v}  ( {Q}_{t} \circ ( \textbf{W}_3 [v_j,\widetilde{e}_{ij}]) ))&\\
    \widetilde{v}_i = \sum_{j=1}^{N}\beta_{ij} v_j&
\end{align}
where ``$[\cdot ,\cdot ]$'' denotes concatenation, $\widetilde{v}_i$ is updated representation of center node and  $v_j$ represents the neighborhood w.r.t. $\widetilde{v}_i$.

\noindent \textbf{Text Knowledge Encoding} 

\noindent For the text knowledge graph, the nodes $S=\{s_i\}^{t}$ ($t$: the round number) are dialogue entities (each Q-A pair in $H_t$), which are extracted by an unbidirectional LSTM and encoded with the concatenation of GloVe \cite{Pennington2014Glove} and ELMo \cite{Peters2018Deep} word embeddings. The edges $R=\{r_{ij}\}^{t\times t}$ are the context dependence between nodes, 
which are provided by semantic dependence encoder via the concatenation of center node $s_i$ and its neighbor $s_j$, i.e. $r_{ij} = [s_i,s_j]$. Based on the text knowledge graph, we further abandon the unrelated information and select query-relevant knowledge for the next Knowledge Storage module. The text knowledge graph is also updated by the \emph{Query-Guided Relation Selection} and \emph{Query-Guided Graph Convolution}, which is similar to the update of vision knowledge graph (just with different inputs). Due to the space limitations, the details of update operation are omitted. Then the updated text knowledge graph is denoted as $\widetilde{S}=\{\widetilde{s}_i\}^{t}$.
% To better capture the context dependency, the edges

\subsection{Knowledge Storage}
\label{msage}

Intuitively, humans are exposed to massive information all the time, while we can merely store impressive information to the knowledge bank. Fortunately, it is easy for a healthy brain to store cross-modal knowledge, since our brain has the amazing ability to keep cross-modal information in diverse modalities simultaneously \cite{paivio1971imagery}. However, how to construct a bridging relation between vision and text knowledge in learning models is particularly important in visual dialogue or even the general AI research. Thanks to the sufficient study of graph neural network (GNN), the structure of GNN can be naturally used as bridges to set up semantic relations between entities cross modalities. To move a further step, we design two cross-modal GNNs (\emph{T2V} and \emph{V2T} mentioned below) to bridge the cross-modal gap and capture the underlying inter-modal semantics when storing knowledge. Knowledge Storage module is presented with orange boxes in Figure \ref{model_pic}, which contains two parts: \emph{Vision Knowledge Storage} and \emph{Text Knowledge Storage}.

\begin{figure}
\centering
\includegraphics[width=8.5cm]{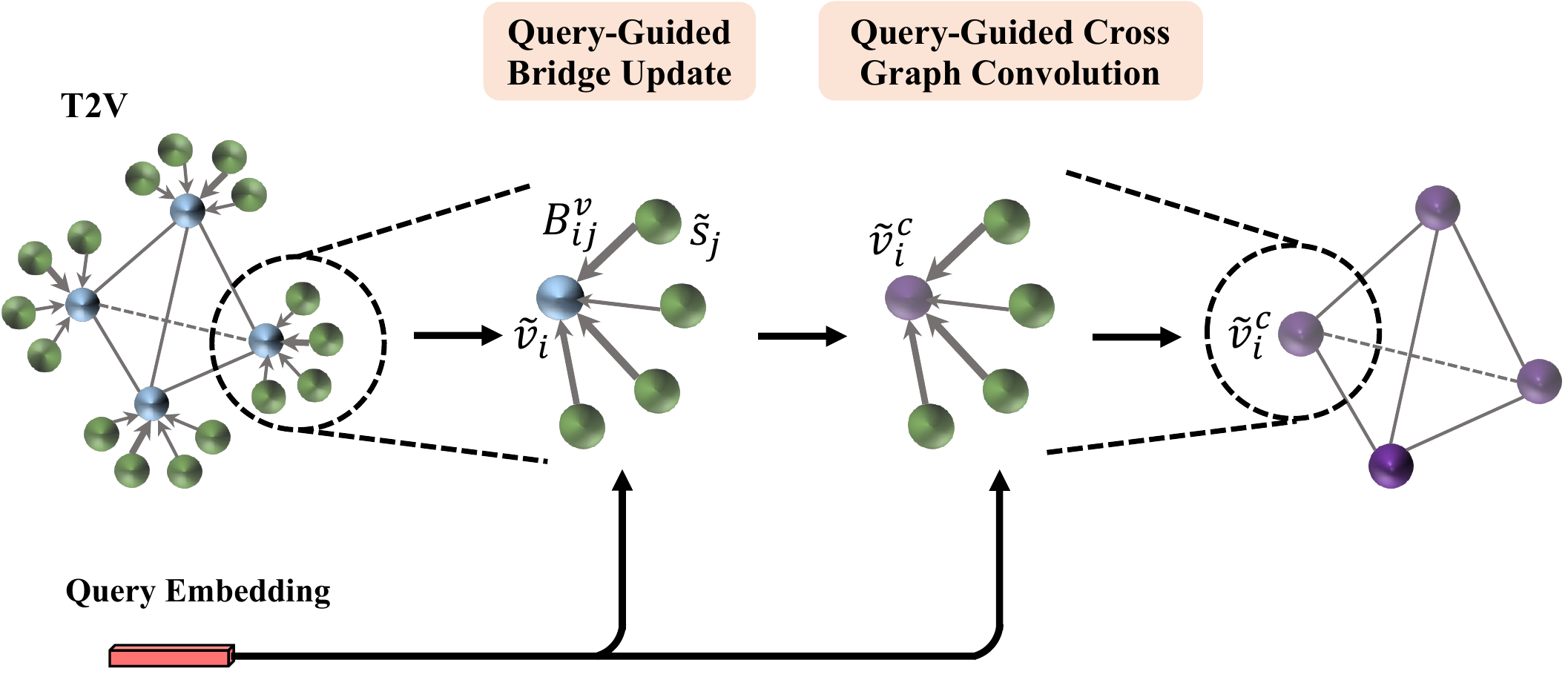}
\caption{The illustration of the cross bridge operation, where T2V is the Text to Vision GNN. We highlight the operation on a specific center node (dotted circle).
The operation of cross-modality relation modeling in text is similar to this but has different inputs.}
\label{bridge-pic}
\end{figure}

\noindent \textbf{Vision Knowledge Storage} 

\noindent For the Text to Vision GNN (T2V, shown in Figure \ref{model_pic}), each intra-modal center node (blue ball) $\widetilde{v}_i$ is connected with all the inter-modal cross nodes (green balls) $\widetilde{S}=\{\widetilde{s}_i\}^{t}$ by $t$ edges $B_{ij}^v$. $B_{ij}^v$ captures underlying semantic dependence from each entity of text knowledge to vision knowledge, which is provided by the cross-modal dependence encoder via the concatenation of $\widetilde{v}_i$ and $\widetilde{s}_{j}$, i.e. $B_{ij}^v = [\widetilde{v}_i,\widetilde{s}_{j}]$. The storage strategy mainly contains two steps: \emph{Cross Bridge} and \emph{Storage}.

\emph{Cross Bridge}: This step aims to update the intra-modal center node with the inter-modal cross node to align different entities from vision and text knowledge. As shown in Figure \ref{bridge-pic}, Cross Bridge contains two parts: \emph{Query-Guided Bridge Update} and \emph{Query-Guided Cross Graph Convolution}.

1) \emph{Query-Guided Bridge Update} aims to capture the underlying semantic dependence between the cross-modal information in local granularity, under the guidance of the query $Q_{t}$:
\begin{align}
\label{br_ww} \gamma _{ij} = softmax(\textbf{W}_b^v  (\textbf{W}_4   Q_{t}  \circ \textbf{W}_5  B_{ij}^v))&\\
\widetilde{B}_{ij}^v = \gamma _{ij}  B_{ij}^v&
\end{align}

2) \emph{Query-Guided Cross Graph Convolution} devotes to introducing inter-modal aligned knowledge via the updated cross-modal bridge as described above:
\begin{align}
    \delta_{ij} = softmax(\textbf{W}_{c}^v  ( {Q}_{t} \circ ( \textbf{W}_6 [\widetilde{s}_j,\widetilde{B}_{ij}^v]) ))&\\
   \label{tsx} \widetilde{v}_i^c = \sum_{j=1}^{t}\delta_{ij} \widetilde{s}_j&
\end{align}
where $\widetilde{v}_i^c$ captures the aligned information of vision knowledge in text domain.

\begin{figure}[t]
\centering
\includegraphics[width=8cm]{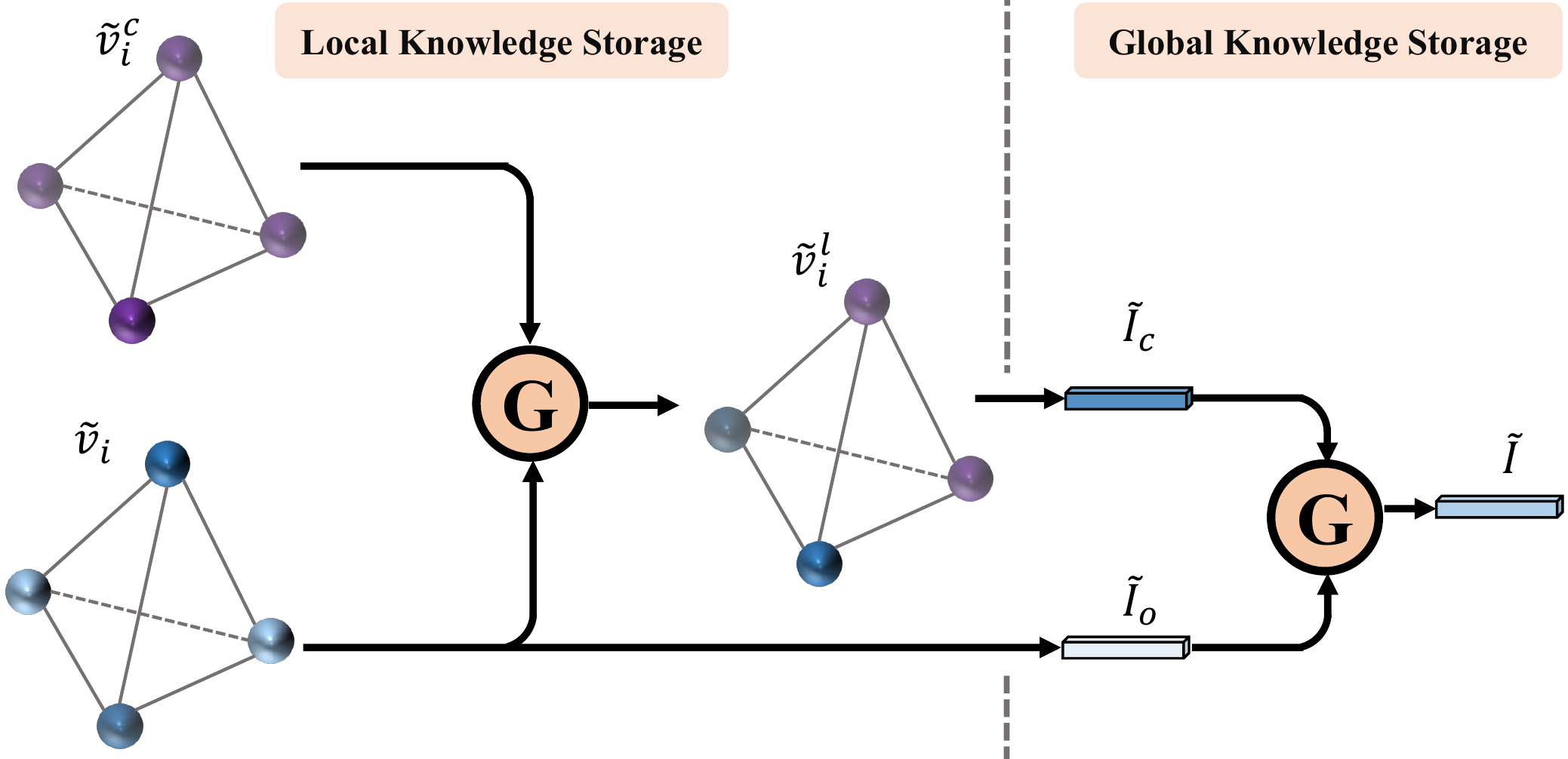}
\caption{The illustration of the Storage operation, where ``G'' represents gate operation given inputs. It mainly contains two parts: local knowledge storage and global knowledge storage. The operation in text knowledge storage is similar to the operation in vision but has different inputs.}
\label{storage-pic}
\end{figure}

\emph{Storage}: Each node in the graph represents the local information of vision knowledge. Storage aims to get the global information for the vision knowledge, guided by the query. As shown in Figure \ref{storage-pic}, Storage contains two parts: \emph{Local Knowledge Storage} and \emph{Global Knowledge Storage}.

1) \emph{Local Knowledge Storage} focuses on the fusion of intra-depend-ence-aware local knowledge $\widetilde{v}_i$ and inter-dependence-aware local knowledge $\widetilde{v}_i^c$. 
%for $\widetilde{v}_i$ is original vision knowledge, whereas $\widetilde{v}_i^c$ contains more information of aligned text knowledge.
We assign a gate operation on these two graph's nodes to obtain the local vision knowledge adaptively:
\begin{align}
    gate_l^v =  \sigma (\textbf{W}_{l}^c [\widetilde{v}_i ,\widetilde{v}_i^c])&\\
   \widetilde{v}_i^l =   \textbf{W}_7 (gate_l^v \circ  [\widetilde{v}_i ,\widetilde{v}_i^c])&
\end{align}

2) \emph{Global Knowledge Storage} targets on the global intra-modal knowledge information generation, since the diverse local knowledge may contain redundant information, as well as human can only store impressive information to the knowledge bank. The original global intra-modal knowledge $\widetilde{I}_o$ is calculated as follows:
\begin{align}
    \eta_i^v =  softmax (\textbf{W}_{e}^v  (Q_t \circ ( \textbf{W}_8  \widetilde{v}_i)))&\\
   \widetilde{I}_o = \sum_{i=1}^{N}\eta_i^v \widetilde{v}_i
\end{align}
The cross global intra-modal knowledge $\widetilde{I}_c$ can be calculated as:
\begin{align}
    \label{att} \mu_i^v =  softmax (\textbf{W}_{a}^v  (Q_t \circ ( \textbf{W}_9  \widetilde{v}_i^l)))&\\
    \widetilde{I}_c = \sum_{i=1}^{N}\mu_i^v \widetilde{v}_i^l&
\end{align}
Then the global vision knowledge is stored by a gate operation:
\begin{align}
    gate_g^v =  \sigma (\textbf{W}_{l}^g [\widetilde{I}_o,\widetilde{I}_c])&\\
   \widetilde{I}  =   \textbf{W}_{10} (gate_g^v \circ  [\widetilde{I}_o,\widetilde{I}_c])&
\end{align}

\noindent \textbf{Text Knowledge Storage}

\noindent For the Vision to Text GNN (V2T, shown in Figure \ref{model_pic}), each intra-modal center node (green ball) $\widetilde{s}_i$ is connected with all the inter-modal cross nodes (blue ball) $\widetilde{V}=\{\widetilde{v}_i\}^{N}$ by $N$ edges $B_{ij}^s$. $B_{ij}^s$ tackles the latent semantic dependence from each entity of vision knowledge to text knowledge, provided by another cross-modal dependence encoder via the concatenation of $\widetilde{s}_i$ and $\widetilde{v}_{j}$, i.e. $B_{ij}^s = [\widetilde{s}_i,\widetilde{v}_{j}]$. Then the storage strategy also contains two steps: \emph{Cross Bridge} and \emph{Storage}, which is similar to the Vision Knowledge Storage (just with different inputs). Due to the space limitations, the details are omitted. After Text Knowledge Storage, the model stores the global text knowledge information $\widetilde{H}_t$.

\subsection{Knowledge Retrieval}

\begin{figure}[t]
\centering
\includegraphics[width=8.5cm]{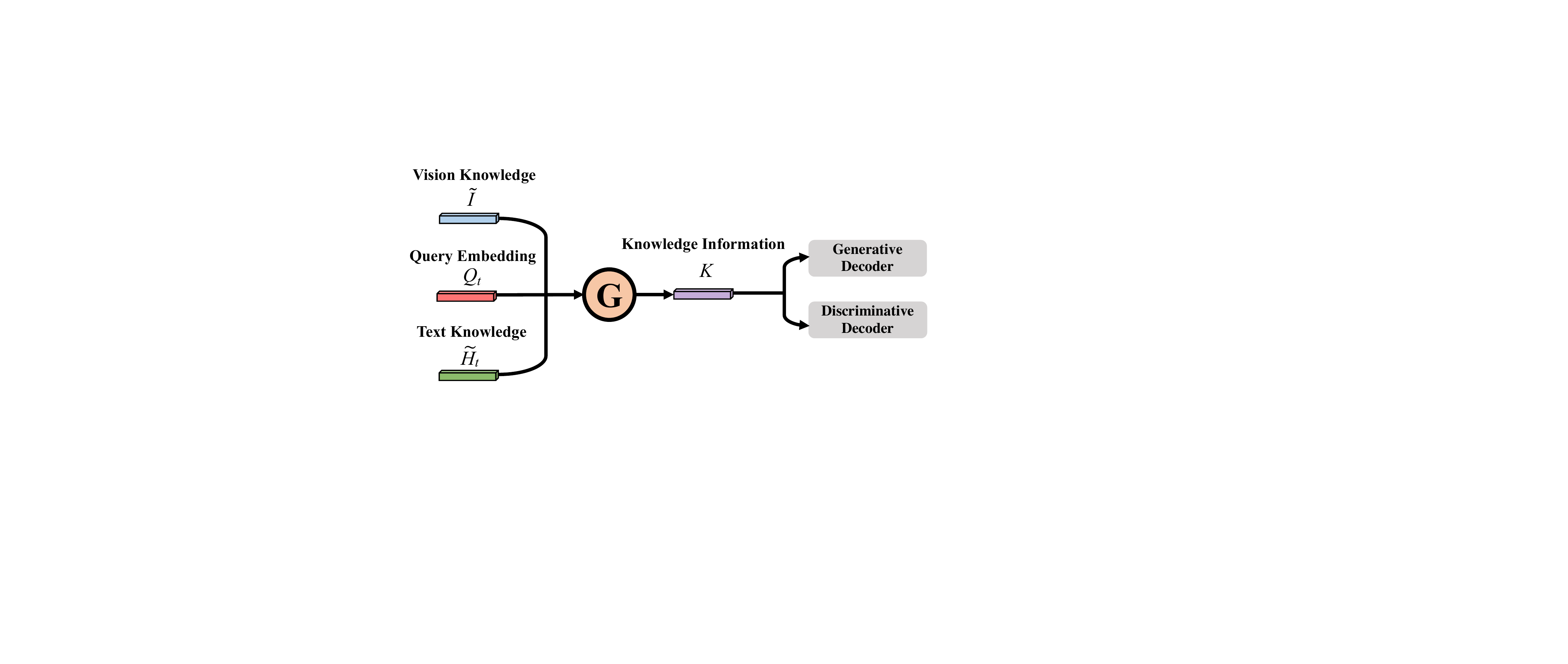}
\caption{The illustration of the Knowledge Retrieval with multi-type decoders, where ``G'' represents gate operation given inputs. Under the awareness of current query, it can adaptively select relevant information to predict the answer. }
\label{re-decoder}
\end{figure}

As mentioned in Section \ref{Introduction}, when responding to the question, the answer clue may come from vision knowledge, text knowledge or both. Previous encoder frameworks, like Late Fusion (LF) \cite{Das2017Visual} and Memory Network (MN) \cite{Das2017Visual}, failed in selecting favorable knowledge adaptively. As shown in Figure \ref{re-decoder}, we assign a gate operation to retrieve valuable knowledge information $K$ from vision knowledge and text knowledge under the awareness of query $Q_t$:
\begin{align}
   \label{gate_mr} gate_r =  \sigma (\textbf{W}_{r} [Q_t, \widetilde{I}, \widetilde{H}_t]) &\\
   K = \textbf{W}_{11} (gate_r \circ  [Q_t, \widetilde{I}, \widetilde{H}_t]) &
\end{align}
Then $K$ is fed to the decoder. There exist two types of decoders in visual dialogue: \emph{discriminative} and \emph{generative}. Discriminative decoder ranks all the answers in the answer candidates $\mathbb{A}$, while generative decoder outputs probability distribution over the vocabulary at each decoding step. More details of these two types decoders are implemented in \cite{Das2017Visual}.

\section{Experiments}
\textbf{Datasets}: We conduct extensive experiments on recently published datasets: VisDial v1.0 \cite{Das2017Visual} and VisDial-Q \cite{ujain2018two}. VisDial v1.0 was collected via two Amazon Mechanical Turk (AMT) subjects chatting about an image. For this dataset, examples are split into \emph{train} (120k), \emph{val} (2k) and \emph{test} (8k), and each dialogue consists of 10 rounds of question-answer pairs on MSCOCO images \cite{lin2014microsoft}. The \emph{train} set contains dialogues regarding COCO-\emph{trainval}, while \emph{val} and \emph{test} set consist of dialogues concerning extra 10k COCO-like images from Flickr. In order to assess the performance of a model asking questions, \citeauthor{ujain2018two} \cite{ujain2018two} first proposed VisDial-Q dataset, which is built upon VisDial v0.9 \cite{Das2017Visual}. VisDial-Q dataset splitting is 80k for \emph{train}, 3k for \emph{val} and 40k as the \emph{test}. %The inputs of VisDial-Q are an image $I$ with its caption $C$, dialogue history till round $t-1$ $H_{t-1}$ and current query $Q^q={Q_t,A_t}$ and the aim is rank a list of 100 candidate questions to predict the next round question $Q_{t+1}$.

\noindent\textbf{Evaluation Metrics}: We follow the metrics in \cite{dai2017detecting} to evaluate response performance. Specifically, the model is required to return a sorting of 100 candidate answer options and evaluated on retrieved metrics: (1) existence of the human response in top $k$ responses, i.e. Recall@$k$ (R@$k, k=1, 5, 10$), (2) mean rank of human response (Mean), (3) mean reciprocal rank (MRR) on VisDial v1.0 and VisDial-Q dataset. For VisDial v1.0, an extra metric, Normalized Discounted Cumulative Gain (NDGG), is involved for a more comprehensive performance study. Lower value for Mean and higher values for all the other metrics are desirable. 

\noindent\textbf{Implementation Details}: For the vision knowledge extractor, we utilize Faster-RCNN \cite{Ren2017Faster} with the ResNet-101 to pick up top 36 object regions (i.e. $N=36$) and produce the 2048-dimension region features. The maximum sentence length of the dialogue history and the current question are set to 20. The hidden state size of LSTM blocks is all set to 512.  The dimension of each edge in the graph is all set to 512. We use Adam \cite{kingma2014adam} optimizer to train our model with 16 epochs and cross entropy loss. We first conduct warm-up strategy which trains the model with initial learning rate 1e-3 and warm-up factor 0.2 for 2 epochs and then utilizes cosine annealing learning strategy with initial learning rate 1e-3  and final learning rate 3.4e-4  for the rest of epochs. The mini-batch size is 15 and the drop ratio is 0.5.

\subsection{Overall Results}

We first conduct experiments on VisDial v1.0 \cite{Das2017Visual}, which is the latest dataset for visual dialogue task. To further evaluate the performance of the proposed model, we conduct experiments on VisDial-Q dataset \cite{ujain2018two}, which aims to predict the generation of the next round question. The results are described as follows.
%  and analysis

\begin{table}[t] 
\centering
\caption{Result comparison on test-standard set of VisDial v1.0 on discriminative method.}
\label{v1-d}
%\begin{center} 
\resizebox{.95\columnwidth}{!}{
\begin{tabular} {L{1.9cm}C{0.6cm}C{0.6cm}C{0.6cm}C{0.69cm}C{0.6cm}C{0.77cm}}
\hline                       
Model & MRR & R\textsl{@}1 & R\textsl{@}5 & R\textsl{@}10 & Mean & NDCG  \\
\hline  
LF \cite{Das2017Visual} &55.42 & 40.95 & 72.45 & 82.83 & 5.95 & 45.31 \\
HRE \cite{Das2017Visual} & 54.16 & 39.93 & 70.47 & 81.50 & 6.41& 45.46 \\  
MN \cite{Das2017Visual} & 55.49 & 40.98 & 72.30 & 83.30 & 5.92 &47.50\\  
%LF-Att\cite{Das2017Visual} &57.07&42.08&74.82&85.05&5.41&40.76\\
%MN-Att\cite{Das2017Visual} & 56.90 & 42.43 & 74.00 & 84.35 & 5.59 & 49.58\\
CorefMN \cite{KotturSatwik2018Visual} & 61.50 & 47.55 & 78.10 & 88.80 & 4.40 &54.70\\
RvA \cite{Niu2018Recursive} &63.03&49.03&80.40&89.83&4.18&55.59\\
DL-61 \cite{guo2019image} &62.20&47.90&80.43&89.95&4.17&57.32\\
DAN \cite{kang2019dual}& 63.20&49.63 &79.75 &89.35 &4.30 & 57.59\\
%ReDAN &64.75& 51.10& 81.73& 90.90& 3.89&57.63\\ 
\hline 
VGNN \cite{Zheng2019Reasoning} & 61.37 &47.33 &77.98&87.83&4.57&52.82\\
FGA \cite{schwartz2019factor}&63.70&49.58&\textbf{80.98}&88.55&4.51&52.10\\
DualVD \cite{jiang2019daulvd}& 63.23 & 49.25 & 80.23 & 89.70 & 4.11 & 56.32\\
CAG \cite{guo2020iterative} & 63.49& 49.85&80.63&90.15&4.11&56.64\\
\hline
\textbf{KBGN (ours)} & \textbf{64.13} & \textbf{50.47} & 80.70 & \textbf{90.16} & \textbf{4.08} & \textbf{57.60} \\
\hline
\end{tabular}  
}

\end{table}

\begin{table}[t] 
\centering
\caption{Result comparison on validation set of VisDial v1.0 on generative method. $\dagger$ Re-trained by \cite{Gan2019Multi}.}%$\dagger$ The models are re-implemented by \cite{Gan2019Multi}.}
\label{v1-g}
%\begin{center} 
\resizebox{.95\columnwidth}{!}{
\begin{tabular} {L{2.2cm}C{0.6cm}C{0.6cm}C{0.6cm}C{0.69cm}C{0.6cm}C{0.77cm}}

\hline                       
Model & MRR & R\textsl{@}1 & R\textsl{@}5 & R\textsl{@}10 & Mean & NDCG  \\
\hline  
MN-G \cite{Das2017Visual} $\dagger$  &47.83&38.01 &57.49&64.08&18.76&56.99\\
HCIAE-G \cite{lu2017best} $\dagger$ &49.07 & 39.72 & 58.23 & 64.73 &  18.43 & 59.70 \\
CoAtt-G \cite{wu2018areyou} $\dagger$ & 49.64 & 40.09 & 59.37 & 65.92 & 17.86 & 59.24 \\ 
ReDAN-G \cite{Gan2019Multi}& 49.60 & 39.95&  59.32 & 65.97 &  17.79  & 59.41 \\
\hline
\textbf{KBGN (ours)} & \textbf{50.05} & \textbf{40.40} & \textbf{60.11} & \textbf{66.82} & \textbf{17.54} & \textbf{60.42}\\
\hline  
\end{tabular}  
}
\end{table}

\setlength{\parskip}{5pt}
\noindent\textbf{Performance on VisDial v1.0}
\setlength{\parskip}{0pt}

\noindent We compare our model KBGN with state-of-the-art discriminative models and generative models on VisDial v1.0. The results are shown in Table \ref{v1-d} and Table \ref{v1-g} respectively. LF focuses on multi-modal information fusion. HRE, MN, CorefMN, RvA and DAN are attention-based models. VGNN, FGA, DualVD and CAG are graph-based models (the second block in Table \ref{v1-d}). DL-61 extends the traditional one-stage solution to a two-stage solution by adding the re-ranking mechanism after traditional methods. 

As shown in Table \ref{v1-d} and Table \ref{v1-g}, KBGN outperforms all the approaches on most metrics whatever in discriminative method or generative method models, especially on R@1 on discriminative method (our model outperforms previous SOTA model CAG by 0.62\% on R@1, while CAG outperforms FGA by 0.27\% on R@1), which demonstrates the superiority of our model. 

The analysis is as follows: 1) The work of DualVD, CAG and VGNN are the most relevant to our model, for VGNN constructs graph on dialogue history and DualVD and CAG utilize scene graph to inference on the image. KBGN boosts the performance on all metrics compared with these three models, which proves the effectiveness of our method. 2) FGA in Table \ref{v1-d} performs a little higher on R@5 than ours, for FGA introduces the information of the candidate answers to the encoder side, while we do not employ answers' information for reasoning. 3) We compare our model with single-step models and traditional generative models. ReDAN \cite{Gan2019Multi} adopts multi-step reasoning and outperforms our model on some metrics. DMRM \cite{chen2019dmrm} and DAM \cite{jiang2020dam} achieve higher performance by designing a more complex generative decoder. HACAN \cite{yang2019making} introduces multi-head attention and two-stage training, achieving comparable results with us. What's more, \citeauthor{qi2019two} \cite{qi2019two} achieves better performance on NDCG by adding NDCG for training loss, while performs worse on other metrics. Applying our model to multi-step reasoning and introducing NDCG to training strategy are insightful future works.

\begin{table}[t] 
\centering
\caption{Result comparison on validation set of VisDial-Q. $\dagger$ We re-train the model on VisDial-Q.}
\label{visualq} 
\begin{tabular}{llllllc} %{L{2.42cm}C{0.68cm}C{0.68cm}C{0.68cm}C{0.8cm}C{0.68cm}}%{llllllX} 
\hline                       
Model & MRR & R\textsl{@}1 & R\textsl{@}5 & R\textsl{@}10 & Mean  \\
\hline  
LF \cite{Das2017Visual} $\dagger$ & 18.45 & 7.80 & 26.12 & 40.78 & 20.42 \\
MN \cite{Das2017Visual} $\dagger$ & 39.83 & 25.80 & 54.76 &69.80 & 9.68\\
SF-QI \cite{ujain2018two} &30.21 &17.38 &42.32 &57.16 &14.03 \\
SF-QIH \cite{ujain2018two} &40.60 & 26.76 & 55.17 & 70.39 & 9.32 \\
VGNN \cite{Zheng2019Reasoning} & 41.26 & 27.15 & \textbf{56.47} & 71.97 & \textbf{8.86}\\

\hline 
\textbf{KBGN (ours)} & \textbf{41.39} & \textbf{27.23} & 56.27 & \textbf{72.01} & 9.19 \\
\hline  
\end{tabular}  
\end{table} 

\setlength{\parskip}{5pt}
\noindent\textbf{Performance on VisDial-Q}
\setlength{\parskip}{0pt}

\noindent We further conduct experiments on VisDial-Q dataset to evaluate the performance of KBGN. The inputs of VisDial-Q are an image $I$ with its caption $C$, dialogue history till round $t-1$, $H_{t} = \{C,(Q_1, A_1),...,(Q_{t-1}, A_{t-1})\}$, and current query $Q_t^q=\{Q_t,A_t\}$. The aim is to rank a list of 100 candidate questions to predict the next round question $Q_{t+1}$. 

Note that we only evaluate discriminative ability of our model on VisDial-Q dataset by convention. The results are shown in Table \ref{visualq}, where SF-QI and SF-QIH are the attention-based models. KBGN outperforms all the approaches on most metrics, except for R@5 and Mean. 

The following reasons may cause the unsuccessful performance on R@5 and Mean: 1) VGNN adopts more complex reasoning operations on  the graph and introduces more complex training algorithms, like EM algorithm, to train the model. The reasoning step for KBGN is not as complex as VGNN and we barely use the traditional training strategy in visual dialogue to train the whole model. More importantly, without the complex reasoning and training strategy like VGNN utilizes, the performance of KBGN is comparable to VGNN's, which also proves the effectiveness of the proposed method. 2) The Knowledge Retrieval module is designed for adaptively selecting information from vision and text knowledge, for the reasoning clues clearly exist in one of the knowledge banks or both. However, the next question prediction may not fit this situation, for the agent can ask any questions within the visual contents, as long as the agent desires. Designing a framework specifically for VisDial-Q task is another future work.
%\end{itemize}

%the performance of KBGN is comparable to state-of-the-art.

\subsection{Ablation Study}

To prove the influence of the essential components of KBGN and the effectiveness of the construction of cross-modal graph neural network, we keep the decoder as the discriminative decoder and then conduct experiments on VisDial v1.0 validation set.

\begin{table}[t] 
\caption{Ablation study of KBGN on validation set of VisDial v1.0 about essential components of KBGN.}
\label{abla-1} 
\centering
\begin{tabular}{lllllllc} %{L{2.42cm}C{0.68cm}C{0.68cm}C{0.68cm}C{0.8cm}C{0.68cm}}%{llllllX} 
\hline                       
Model & MRR & R\textsl{@}1 & R\textsl{@}5 & R\textsl{@}10 & Mean & NDCG \\
\hline  
VTA   & 63.77 & 49.50 & 80.33 & 90.05 & 4.25 & 56.50\\
VETA & 64.23 & 50.19 & 80.62 & 90.17 & 4.17 & 57.19\\
VETE & 64.43 & 50.91 & 80.96 & 90.21 & 4.12 & 57.60 \\
VT2V & 64.52 & 50.99 & 81.14 & 90.35 & 4.07 & 58.11\\
TV2T & 64.60 & 51.07 & 81.33 & 90.46 & 4.03 & 58.53\\

\hline 
\textbf{KBGN} & \textbf{64.86} & \textbf{51.37} & \textbf{81.71} & \textbf{90.54} & \textbf{4.00} & \textbf{59.08}  \\
\hline  
\end{tabular}  
\end{table} 

\setlength{\parskip}{5pt}
\noindent\textbf{The Effectiveness of Essential Components}
\setlength{\parskip}{0pt}

\noindent The whole architecture mainly incorporates three essential components: intra-knowledge graph, cross-modal graph and knowledge retrieval gate. We consider the following ablation models to verify the influence of each component in vision knowledge and text knowledge:

(1) \textbf{VTA}: this is our baseline model. It utilizes the question to attend each entity of vision knowledge and text knowledge to get global vision knowledge and global text knowledge. Then the Late Fusion framework \cite{Das2017Visual} is adopted to fuse the information of question, global vision knowledge and global text knowledge. 

(2) \textbf{VETA}: compared with VTA, this model applies vision knowledge graph to encode and update vision knowledge. 

(3) \textbf{VETE}: compared with VETA, this model applies text knowledge graph to encode and update text knowledge. 

(4) \textbf{VT2V}: compared with VETE, this model utilizes Vision to Text GNN to capture the latent semantic dependence. 

(5) \textbf{TV2T}: compared with VT2V, this model utilizes Text to Vision GNN to grip implicit semantic dependence. 

(6) \textbf{KBGN}: this is our comprehensive model, which further uses Knowledge Retrieval module to select question-relevant knowledge. 

The results are shown in Table \ref{abla-1}, the key observations and analysis are as follows: 1) \textbf{VETA} considers vision knowledge encoding by constructing graph neural network on vision entities. \textbf{VETE} focuses on the semantic dependence between each entity in text knowledge. After introducing graph neural network to vision knowledge and text knowledge, it increases by 0.69\% and 0.41\% on NDCG respectively. This illustrates the utility of the encoding and updating strategy on vision and text knowledge via Knowledge Encoding module. 2) \textbf{VT2V} devotes to the construction of text to vision cross-modal bridge. \textbf{TV2T} pays more attention to the connection of vision knowledge to text knowledge. After the consideration of involving Vision to Text GNN (V2T) and Text to Vision GNN (T2V) to Knowledge Storage module, it increases by 0.51\% and 0.42\% on NDCG respectively, which proves the advantages of the construction of cross-modal bridge when capturing the underlying semantic dependence in cross-modal information. 3) The models mentioned above directly concatenate the cross-modal information to fusion, while \textbf{KBGN} moves a further step to retrieve reasoning clues via the gate operation. It further increases the performance on NDCG by 0.55\%, which confirms the effectiveness of the information selection mode adopted by Knowledge Retrieval. How to construct a more complex and adaptive fusion strategy is a bright future work.

% information from vision knowledge bank and text knowledge bank adaptively. It further increases the performance on NDCG by 0.55\%, which confirms the effectiveness of the information selection mode adopted by Knowledge Retrieval.

%\end{itemize}

\setlength{\parskip}{5pt}
\noindent\textbf{The Effectiveness of Cross-modal Graph Neural Network}
\setlength{\parskip}{0pt}

\begin{table}[t] 
\centering
\caption{Ablation study of KBGN on  VisDial v1.0 validation set about the construction of cross-modal GNN.}
\label{abla-2} 

\begin{tabular}{lllllllc} %{L{2.42cm}C{0.68cm}C{0.68cm}C{0.68cm}C{0.8cm}C{0.68cm}}%{llllllX} 
\hline                       
Model & MRR & R\textsl{@}1 & R\textsl{@}5 & R\textsl{@}10 & Mean & NDCG \\
\hline  
V-NoRel & 64.64 &  50.94 & 81.33  & 90.20 & 4.21 & 58.65\\
V-RRel & 64.71 & 51.26 & 81.54 & 90.35  & 4.11 & 58.84\\
T-NoRel &64.59 & 50.96 & 81.28 & 90.26 & 4.20 & 58.71\\
T-RRel & 64.70 & 51.30 & 81.55 & 90.33 & 4.14  & 58.86\\
\hline 
\textbf{KBGN} & \textbf{64.86} & \textbf{51.37} & \textbf{81.71} & \textbf{90.54} & \textbf{4.00} & \textbf{59.08}  \\
\hline  
\end{tabular}  
\end{table}

\noindent After the examination of essential components of KBGN, we will further evaluate the method that we use in the construction of cross-modal graph neural network. To our best knowledge, we are the first to apply graph structure to build knowledge bridges in visual dialogue. We choose the following ablation models to prove the effectiveness of cross-modal graph neural network:

(1) \textbf{V-NoRel}: this model replaces relation embeddings with unlabeled edges on Text to Vision GNN and the convolution is computed as inter-modal attention.

(2) \textbf{V-RRel}: this model adopts random initialization to initialize the relation embeddings on Text to Vision GNN.

(3) \textbf{T-NoRel}: similar to V-NoRel, this model replaces relation embeddings with unlabeled edges on Vision to Text GNN and the convolution is computed as inter-modal attention.

(4) \textbf{T-RRel}: similar to V-RRel, this model adopts random initialization to initialize the relation embeddings on Vision to Text GNN.

(5) \textbf{KBGN}: this is our full model, which utilizes the concatenation of intra-modal center nodes and inter-modal cross nodes as the initial relation embeddings (implemented in Section \ref{msage}).

The results are shown in Table \ref{abla-2}, the key observations and analysis are as follows: 
1) \textbf{V-NoRel} and \textbf{T-NoRel} bridge the cross-modal knowledge information without relation dependence. Compared with KBGN, the performance of V-NoRel and T-NoRel on NDCG decreases by 0.43\% and 0.37\% respectively, which proves that the relation dependence exists in cross-modal information. 2) \textbf{V-RRel} and \textbf{T-RRel} take relation dependence into consideration and randomly initialize the relation embeddings. It increases by 0.19\% and 0.15\% on NDCG respectively, while sill has 0.24\% and 0.22\% gap to KBGN separately. 3) \textbf{KBGN} further views the combination of cross-modal information as initial relation dependence, which performs best on all the metrics among the models in Table \ref{abla-2}. It demonstrates that the method, adopted by us, has dominant advantages in bridging the cross-modal gap in visual dialogue. What's more, we also vote for more powerful construction strategy to the cross-modal graph neural network in visual dialogue.
\begin{figure*}[t]

\centering
\includegraphics[width=18cm]{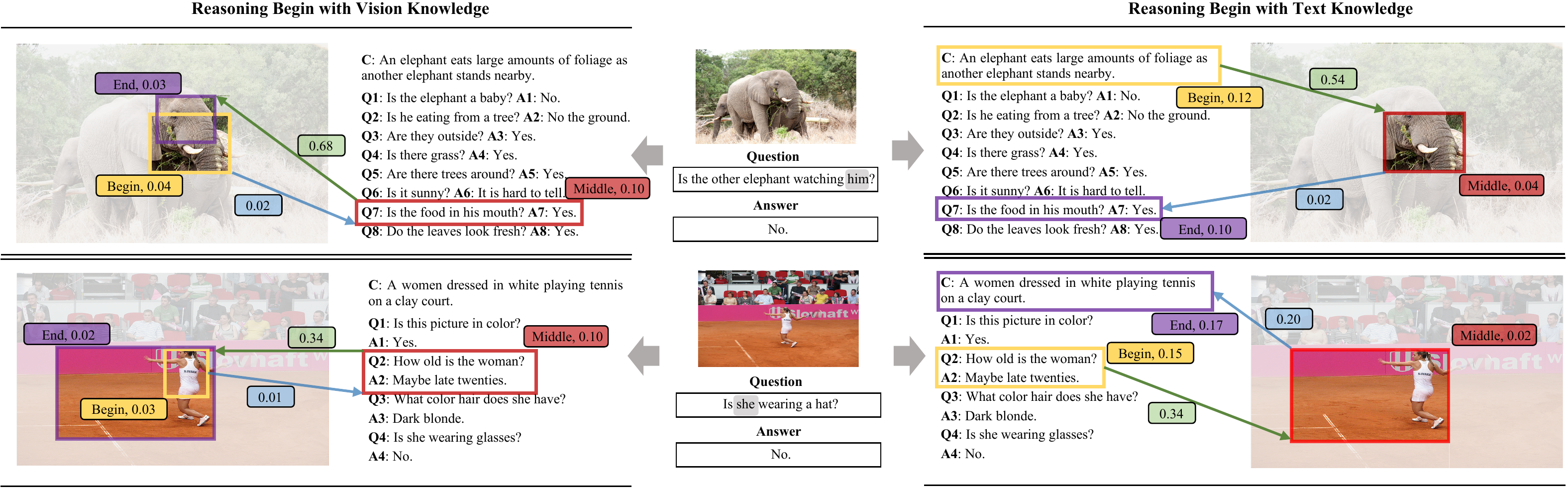}
\caption{Visualization for cross-modal bridge reasoning. 
%, the model retrieves reasoning clues in two aspects: Reasoning begin with Vision Knowledge and Reasoning Begin with Text Knowledge. Take Reasoning Begin with Vision Knowledge 
For example, we first highlight the most relevant vision object (yellow box) according to attention weights of each object ($\mu_i^v$ in Eq. \ref{att}). Then we can find the relevant text region (red box) with the top one attended V-T (vision to text) relationships ($\gamma _{ij}$ in Eq. \ref{br_ww}). In the end, we underline the vision object (purple box) with the top one attended T-V relationships.}

\label{visual-bridge}
\end{figure*}
\begin{figure}[t]
\centering
\includegraphics[width=8cm]{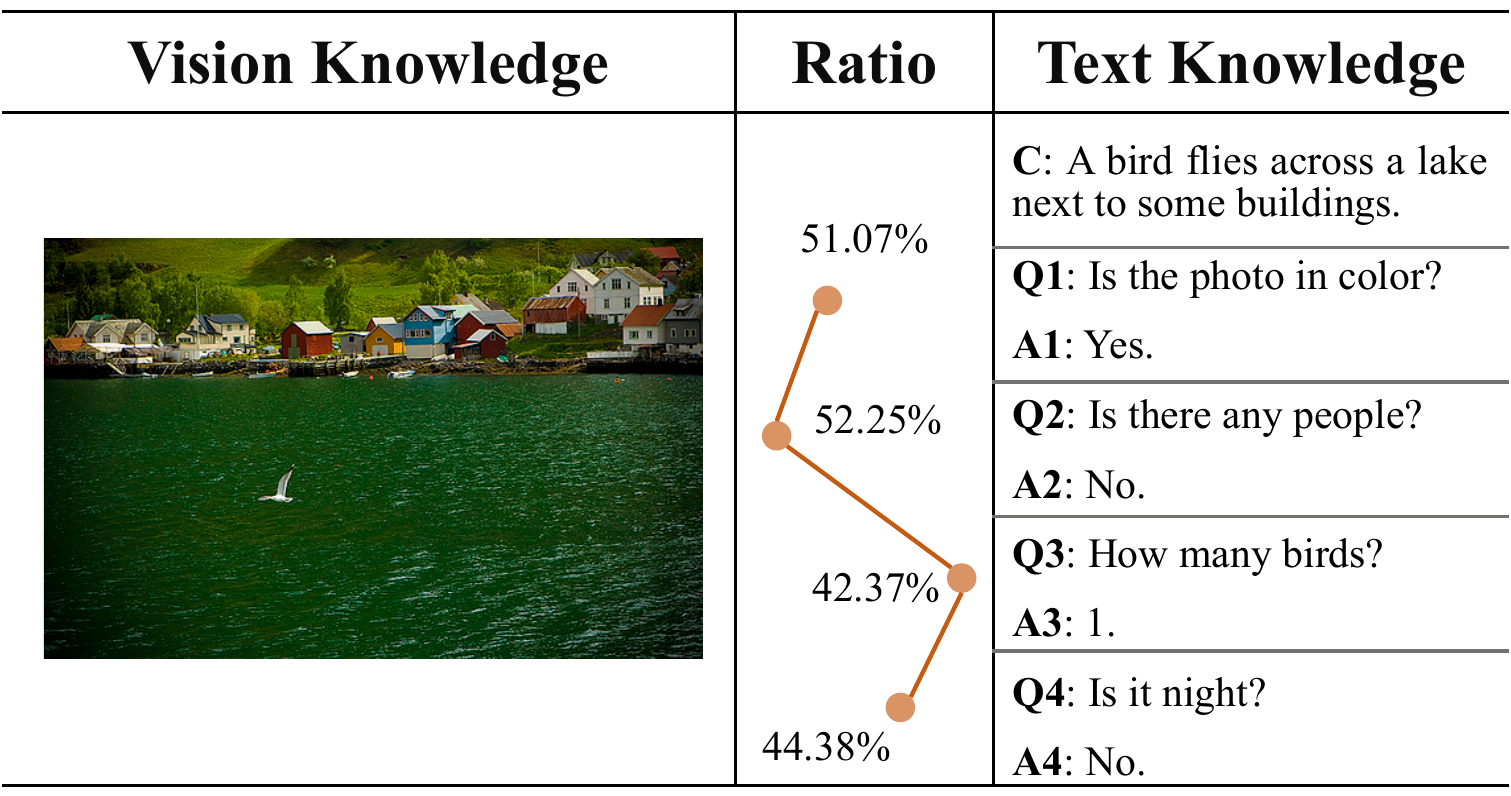}
\caption{Visualization for adaptive knowledge retrieval reasoning. The orange polyline represents the ratio of total gate values for vision knowledge (Eq. \ref{gate_mr}, abandon query gate) when answering the Q1, ..., Q4. }
\label{ratio-pic}
\end{figure}

% \begin{itemize}

% \end{itemize}

\subsection{Qualitative Analysis}

We further review the reasoning evidence by visualizing the process of cross-modal bridge reasoning and knowledge information retrieval on VisDial v1.0 validation set on discriminative method.

\noindent\textbf{Cross-modal Bridge Reasoning}

\noindent To figure out how the cross-modal graph network works, we visualize the cross-modal bridge reasoning process. As shown in Figure \ref{visual-bridge}, there are two examples with two reasoning clues for answering the question. Take the \emph{Reasoning Begin with Text Knowledge} for example to analyze: When answering `` \emph{Is the other elephant watching him?}'', to determine what is `` \emph{him}'' referring to, the model focuses on the caption $C$ to get the reasoning clue  `` \emph{an elephant}'' according to intra-modal semantic dependence. Since the cross-modal bridge exists in cross-modal knowledge information, the model accurately finds the corresponding representation (the visual object with red box in the first example) in the vision knowledge. Furthermore, the model links the vision knowledge with \emph{Q7A7} via cross-modal bridge, for \emph{Q7A7} is the description of vision knowledge in text domain. And then it resolves `` \emph{him}'' on the elephant that is eating grass, which also reveals that the cross-modal relation dependence exists in cross-modal information. Similar observation exists in the second example, illustrating that the cross-modal graph network proposed by us can bridge the cross-modal gap successfully, as well as capture the underlying semantic dependence between each entity of vision and text knowledge. %More visualization cases are implemented in supplementary material.
% also contains the reasoning clue `` \emph{his}''.

\noindent\textbf{Adaptive Knowledge Retrieval Reasoning}
 
 \noindent Another advantage of KBGN is that it selects the knowledge  from different modalities adaptively, under the awareness of questions. As shown in Figure \ref{ratio-pic}, the ratio polyline reveals the information selection mode of KBGN when facing diverse questions: Confronted with ``\emph{Is the photo in color?}'' and ``\emph{Is there any people?}'', the model focuses more on the vision knowledge to retrieve ``\emph{color}'' and ``\emph{people}'' in the photo. Whereas answering ``\emph{How many birds?}'', the model pays more attention to the text knowledge, going back to the caption $C$ to get the reasoning clue ``\emph{A bird}''. Moreover, to define what is ``\emph{it}'' in \emph{Q4}, the model selectively concentrates on the text knowledge and grounds the reference ``\emph{it}'' to the ``\emph{photo}'' in \emph{Q1}, revealing that the model can adaptively select information from vision and text knowledge. This observation also exists in lots of other cases from the dataset.

\section{Conclusion}

 In this paper, we propose a novel model for visual dialogue by capturing underlying semantic dependence, as well as retrieving advisable information adaptively in two modalities. We applied graph neural network (GNN) in modeling the relations between the visual dialogue cross-modal information in fine granularity. Experimental results on two benchmark large-scale datasets illustrate the superiority of our new proposed model. More importantly, the reasoning clues can be clearly seen by utilizing the proposed model. Last but not the least, we vote for other more powerful construction and training strategy to the cross-modal graph neural network, which is also our future work.

\bibliographystyle{ACM-Reference-Format}
\bibliography{sample-base}

\appendix









\end{document}